\documentclass[letterpaper, 10 pt, conference]{ieeeconf}  

\IEEEoverridecommandlockouts                              

\usepackage{graphicx} 
\usepackage{amsmath}
\usepackage{amssymb}  
\usepackage{booktabs}
\usepackage{multirow}

\usepackage{subcaption}

\title{\LARGE \bf
Imagine then Verify: Affordance-Targeted Active Perception for Task-Oriented Grasping in Cluttered Scenes
}

\author{Jingzhi Cui, Xuefeng Liu, Feng Han, Xinyu Liu, Wei Chen, Jianwei Niu
\thanks{Jingzhi Cui, Xuefeng Liu, Feng Han, Wei Chen and Jianwei Niu are with the School of Computer Science and Engineering, Beihang University, Beijing, China.
Xinyu Liu is with the School of Artificial Intelligence, Beihang University, Beijing, China.
}%
}

\begin{document}
\bstctlcite{myBSTcontrol}

\maketitle
\thispagestyle{empty}
\pagestyle{empty}

\begin{abstract}

Task-oriented grasping (TOG) requires robots to grasp functional parts of objects (e.g., the handle of a mug for pouring), yet these affordance regions are frequently occluded in cluttered scenes.
Active perception via next-best-view (NBV) planning can resolve such occlusions by moving the camera for more informative observations.
However, existing NBV methods typically optimize viewpoints for grasping the target object as a whole without distinguishing which part is task-relevant. 
A naive adaptation, fully scanning the target object before predicting the affordance, wastes most of the viewpoint budget on task-irrelevant surfaces (e.g., the mug body for pouring).
To address this, we propose ATAP, an Affordance-Targeted Active Perception framework that shifts viewpoint planning from exhaustive target scanning to targeted affordance verification.
ATAP hypothesizes the occluded target geometry via a generative shape prior and predicts the affordance distribution over the imagined complete surface. 
In cluttered scenes, severe occlusion can make the location of the hidden affordance ambiguous, leaving multiple locations plausible given the partial observation.
ATAP therefore introduces an uncertainty-aware viewpoint planner that jointly optimizes expected entropy reduction over these competing hypotheses and expected affordance verification gain from real observations.
This process iterates until the affordance is sufficiently verified for grasp execution.
Experiments in simulation and real-world cluttered scenes show that ATAP substantially improves the functional grasp success rate over fixed-view TOG baselines, and outperforms reconstruction-based active perception with over 57\% fewer NBV steps.

\end{abstract}

\section{INTRODUCTION}

Grasping is a fundamental step in robotic manipulation, but the success of downstream tasks often depends not on whether an object is grasped, but on how it is grasped \cite{catgrasp}. 
When humans grasp an object, we intuitively choose the task-relevant functional part, e.g., the handle of a mug for pouring.
This requirement is formalized as task-oriented grasping (TOG) \cite{tog, foundationgrasp}. Unlike general grasping that aims for any stable grasp on an object \cite{graspnet-1b, anygrasp}, TOG requires the robot to contact the functional part (the affordance region) according to the task instruction.
Existing TOG methods, whether through training-based affordance transfer \cite{foundationgrasp, graspgpt, rtagrasp} or zero-shot VLM reasoning \cite{affordgrasp, ovalgrasp, thinkgrasp}, implicitly assume that the affordance region is visible from the current viewpoint. 
However, in cluttered environments, these functional parts are frequently occluded by surrounding objects or hidden due to unfavorable object orientations, causing affordance reasoning failures and unsuccessful TOG.

\begin{figure}[t]
\centering
\includegraphics[width=1.0\linewidth]{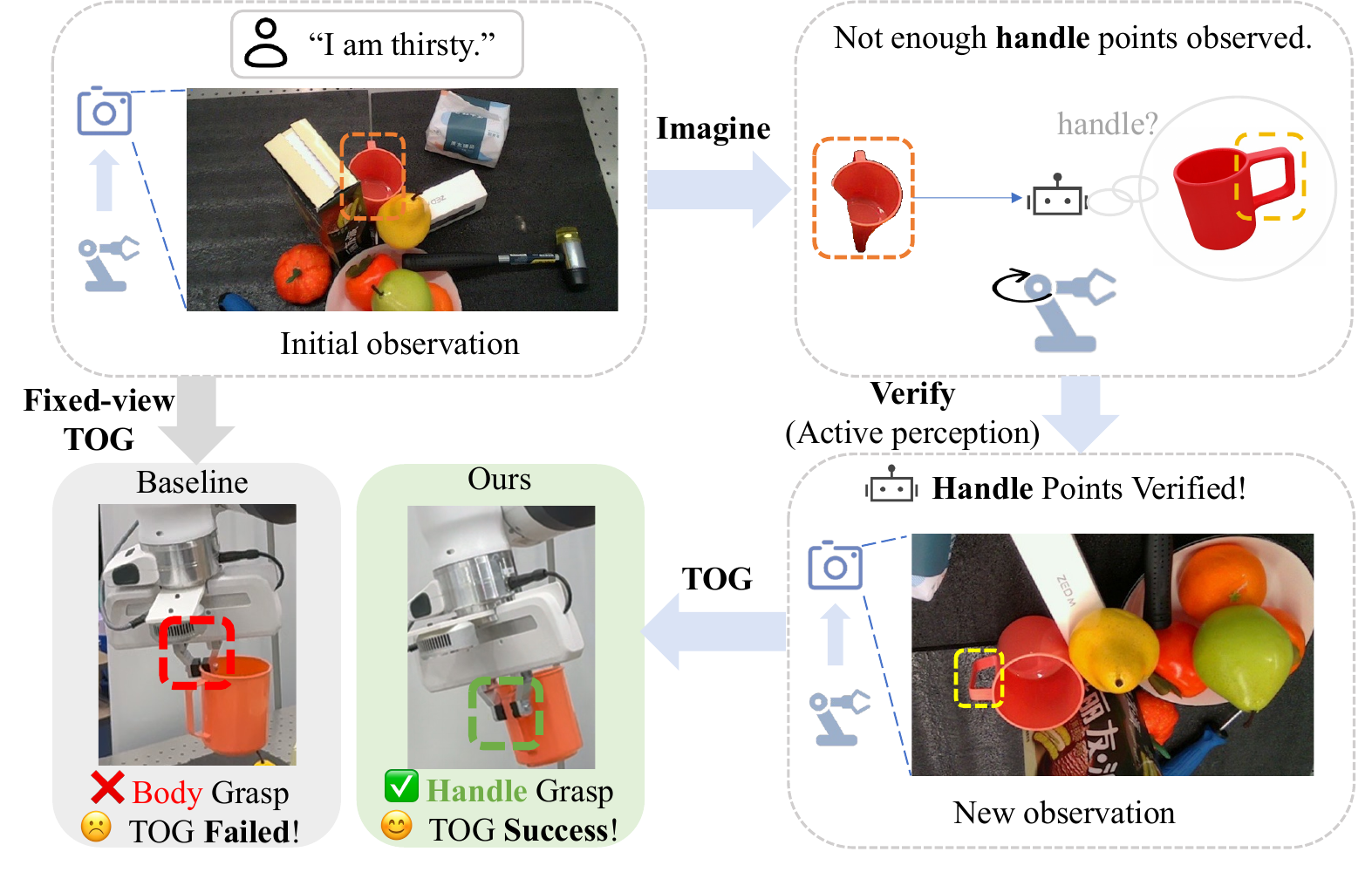}
\caption{
Motivation of \textbf{ATAP}. 
When the task-relevant affordance is occluded, fixed-view TOG tends to grasp the visible body and fails the pouring task. ATAP hypothesizes the hidden affordance from imagined geometry and actively verifies it from a new viewpoint before executing the TOG.}
\label{fig:initial}
\end{figure}

Active perception via Next-Best-View (NBV) planning offers a principled remedy for occlusion by iteratively repositioning the camera for more informative observations \cite{breyer2022, ace-nbv, apeg, viso-grasp}. 
However, existing NBV objectives typically optimize for either geometric reconstruction \cite{breyer2022}, grasp quality \cite{ace-nbv}, or target visibility \cite{apeg, viso-grasp}. 
None of these objectives is explicitly aligned with the task-specific functional affordance required for TOG. 
The most direct way to extend such methods to TOG is a reconstruct-then-reason pipeline that first reconstructs the entire target object and then predicts its affordance. 
This approach is inherently inefficient, as it expends most viewpoint budget on task-irrelevant surfaces (e.g., the body of a mug when only the handle matters for pouring), or terminates once the object is broadly covered while the affordance region remains hidden.

Our key insight is that, for TOG, the affordance region itself requires high-fidelity observation; complete target reconstruction is unnecessary. However, this insight exposes a circular dependency: directing the camera toward the affordance requires knowing its location, yet under occlusion this is precisely the information that must be recovered. 
To address this, we propose ATAP, an Affordance-Targeted Active Perception framework that shifts viewpoint planning from exhaustive object scanning to targeted affordance verification, as illustrated in Fig.~\ref{fig:initial}. 
Given a language instruction and an initial observation, ATAP first hypothesizes the occluded target geometry via a generative shape prior and predicts the task-relevant affordance distribution over the imagined complete surface. 
However, under sparse initial observations the shape prior can yield multiple plausible affordance placements, particularly for symmetric objects where the functional region could lie at several orientations.
To handle this ambiguity, ATAP introduces an uncertainty-aware viewpoint planner that jointly maximizes expected entropy reduction over competing affordance-location hypotheses and expected verification gain from real observations.
This hypothesize-then-verify loop iterates until the affordance is sufficiently verified and the remaining location ambiguity is low, progressively grounding the imagined geometry with real observations before TOG execution.

In summary, our main contributions are as follows: 
\begin{itemize}
    \item We propose ATAP, a closed-loop active perception framework for task-oriented grasping under occlusion. By leveraging generative shape priors, ATAP follows a hypothesize-then-verify strategy that shifts viewpoint planning from exhaustive target reconstruction to targeted affordance verification.
    \item We introduce an affordance-targeted NBV planner that maintains multiple affordance-location hypotheses under ambiguous registration and selects views by jointly optimizing expected affordance verification gain and affordance-location disambiguation.
    \item Simulation experiments and real-robot validation demonstrate that ATAP substantially improves functional grasp success over fixed-view TOG baselines, and outperforms reconstruction-based active perception with over 57\% fewer NBV steps.
\end{itemize}

\section{Related Work}

\subsection{Task-oriented Grasping}

Unlike general grasping, which seeks any stable grasp on an object \cite{graspnet-1b, anygrasp, contactgrasp-net}, task-oriented grasping \cite{tog} requires robots to grasp the functional part according to the task instruction to enable downstream manipulation.
Early TOG methods learn task-grasp mappings from large-scale annotated datasets through semantic knowledge \cite{tog}, grasp-action embeddings \cite{Gater}, or part affordance grounding \cite{partAffordance}, but are limited to closed-set categories.
Recent work improves generalization through LLM-augmented affordance transfer \cite{foundationgrasp, graspgpt, rtagrasp} and zero-shot VLM reasoning \cite{affordgrasp, ovalgrasp, thinkgrasp, shapegrasp}. However, these methods have been validated primarily in single-object or lightly cluttered settings, and implicitly assume the affordance region is visible from the current viewpoint. 
Although AffordGrasp \cite{affordgrasp} and OSTG \cite{ostg} extend TOG to cluttered scenes, they do not address occlusion of the affordance itself. 
In practice, functional parts are frequently occluded by surrounding objects or hidden due to viewing angles, causing TOG failure. 
More recently, TOSC \cite{tosc} addresses partial observability for dexterous hand grasping by completing only the task-relevant contact regions and directly synthesizing grasps.
In contrast, ATAP imagines the full object geometry to support both affordance reasoning and viewpoint planning, and executes a grasp after the affordance has been sufficiently corroborated through active perception.  


\begin{figure*}[thpb]
\centering
\includegraphics[width=0.99\linewidth]{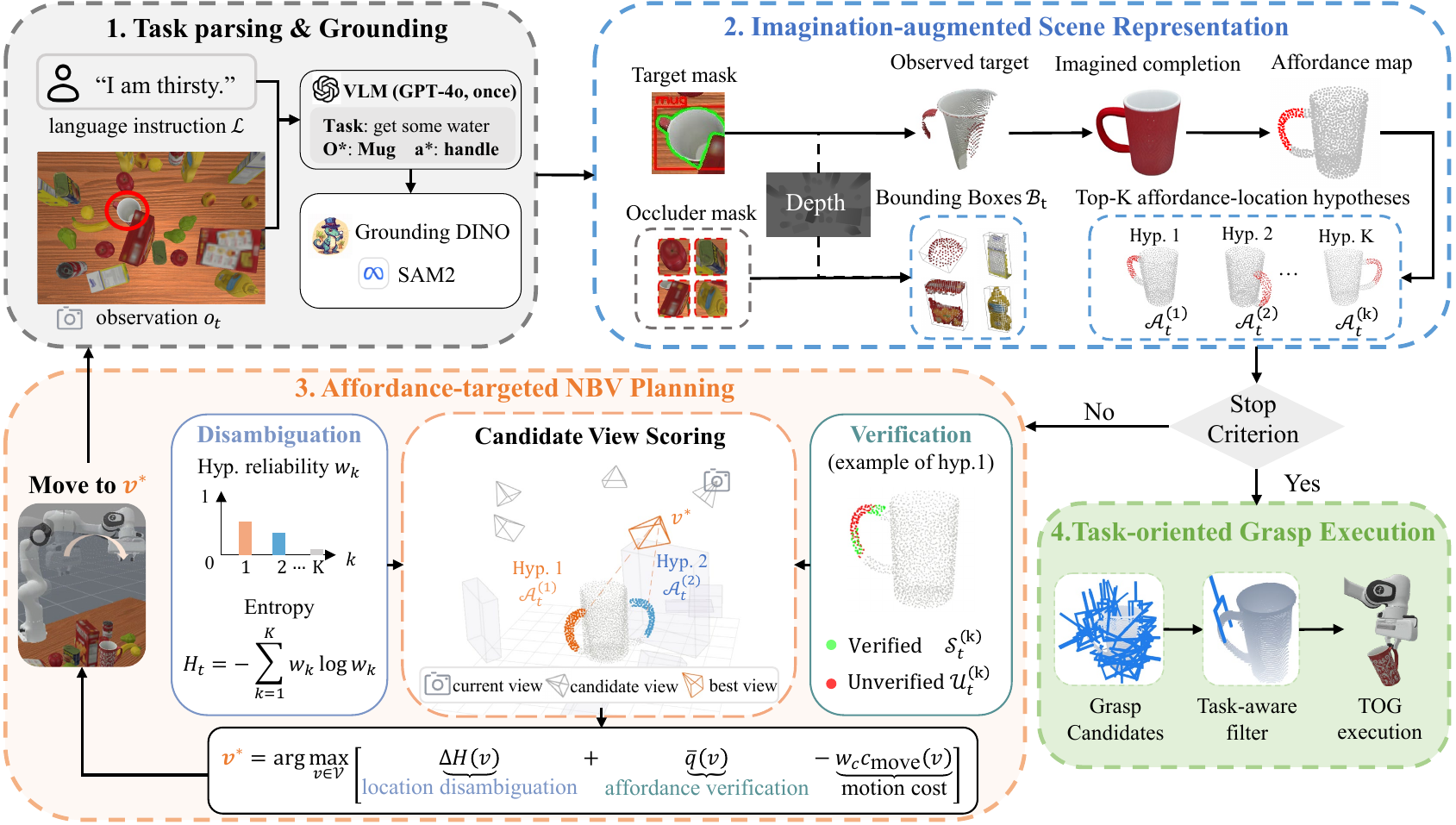}
\caption{
Overview of ATAP.
Given a language instruction and an RGB-D observation, ATAP completes the target geometry, predicts task-relevant affordances, and maintains multiple affordance-location hypotheses.
The NBV planner selects target-centered views by jointly considering affordance verification, hypothesis disambiguation, and motion cost.
The loop continues until the affordance is sufficiently verified and the remaining ambiguity is low, or the viewpoint budget is exhausted, after which the verified affordance guides task-oriented grasping.
}
\label{fig:pipeline}
\end{figure*}

\subsection{Active Perception for Grasping}

Active perception through Next-Best-View (NBV) planning provides a principled approach to resolving occlusion in grasping by iteratively selecting informative viewpoints \cite{breyer2022, ace-nbv, active-ngf}. 
Traditional methods primarily formulate NBV as a reconstruction-driven or grasp-quality-driven problem, leveraging volumetric information gain from ray casting \cite{breyer2022}, grasp imagery at novel views \cite{ace-nbv}, or neural graspness fields \cite{active-ngf}. 
ACE-NBV~\cite{ace-nbv} considers grasp affordance in view selection, but targets generic graspability rather than task-specific functional regions.
%
Recent semantic-guided methods use VLMs or semantic information gain for open-vocabulary target search and grasping~\cite{ap-vlm,viso-grasp, apeg}. However, they still optimize object-level visibility rather than task-relevant affordance verification.
While effective for target localization, these methods treat all occluded geometry of the target as equally informative, overlooking the specific requirements of downstream manipulation.
For TOG, this task-agnostic exploration wastes the viewpoint budget on task-irrelevant surfaces, or terminates prematurely while the affordance region remains hidden. 
To bridge this gap, our framework hypothesizes the affordance distribution via a generative shape prior and introduces an uncertainty-aware viewpoint planner that jointly optimizes affordance verification and affordance-location disambiguation.

\section{Problem Formulation}

We consider a robotic arm with a wrist-mounted RGB-D camera and a parallel-jaw gripper operating over a cluttered tabletop containing $m$ arbitrarily placed objects $\mathcal{O}=\{O_1,\dots,O_m\}$. 
Given a language task instruction \(\mathcal L\) (e.g., ``pour some water"), the robot is required to identify the task-relevant target object $O^*\in\mathcal{O}$ and execute a 6-DoF grasp on its functional part---the \emph{affordance region} $\mathcal{A}^*$---that enables the intended manipulation task. 
A trial is considered successful if the robot lifts $O^*$ stably and the grasp contact falls within~$\mathcal{A}^*$.

At each step $t$, the robot receives observation 
$\mathbf{o}_t=(\mathbf{I}_t,\;\mathbf{D}_t,\;\mathbf{T}_{t}^{ee})$, 
including the RGB frame 
$\mathbf{I}_t\!\in\!\mathbb{R}^{H\times W\times 3}$,
the depth map $\mathbf{D}_t\!\in\!\mathbb{R}^{H\times W}$, and the end-effector pose $\mathbf{T}_{t}^{\mathrm{ee}}\!\in\!SE(3)$. 
Based on the accumulated observations $\mathbf{o}_{0:t}$ and \(\mathcal L\), the system evaluates whether $\mathcal{A}^*$ is sufficiently observable to execute a stable task-oriented grasp. 
A viewing policy $\pi_v\!:(\mathbf{o}_{0:t},\mathcal L)\to\!v_{t+1}\!\in\!SE(3)$ iteratively repositions the camera to expose the occluded affordance geometry. The perception loop continues until $\mathcal{A}^*$ is sufficiently observed or the viewpoint budget $T$ is exhausted, at which point a grasping policy $\pi_g\!:(\mathbf{o}_{0:t},\mathcal{L})\!\to\!g$ is invoked to produce the final 6-DoF grasp pose and gripper width.

\section{Method}

\subsection{System Overview}

As illustrated in Fig.~\ref{fig:pipeline}, ATAP follows a hypothesize-then-verify strategy for task-oriented grasping under affordance occlusion.
Given a task instruction and partial target observations, it first imagines the complete target geometry and predicts task-relevant affordances on the completed surface.
To account for ambiguity in aligning the imagined geometry with partial observations, ATAP retains multiple plausible registrations as competing affordance-location hypotheses.
The NBV planner then selects viewpoints that jointly disambiguate these hypotheses and verify the predicted affordance against real depth observations.
Each new observation updates the scene representation, and this closed-loop process continues until the affordance is sufficiently verified and the remaining hypothesis ambiguity is low, or the viewpoint budget is exhausted.
Finally, the available verified affordance guides grasp selection.

\subsection{Imagination-Augmented Scene Representation}
\label{sec:scene}
Since the task-relevant affordance may be occluded from the initial view, planning toward it requires reasoning about unobserved geometry.
ATAP addresses this with a hybrid scene representation: the target object is represented by a completed, affordance-annotated geometry that captures possible functional-region locations, while surrounding clutter is approximated with coarse 3D
bounding boxes. 
This avoids exhaustive scene reconstruction and focuses the NBV planner on the affordance region.

\textbf{Task Parsing and Target Grounding.}
At episode start, we query a VLM~\cite{openai2024gpt4ocard} once with the task instruction $\mathcal{L}$ and the initial RGB observation $I_0$ to identify the target object $O^*$ and affordance type $a^*$.
The target name guides open-vocabulary grounding, and $a^*$ serves as the language query for subsequent affordance prediction.
Given $O^*$, Grounding DINO~\cite{GroundingDino} and SAM2~\cite{ravi2024sam2} produce the target mask $\mathcal{M}_t$ from $I_t$. 
The masked depth is back-projected to the robot base frame to obtain the observed target cloud
\begin{equation}
  \mathcal{P}^{\text{obs}}_t = \bigl\{
    \mathbf{T}^{\text{cam}}_t \cdot \mathbf{K}^{-1}[u,v,1]^\top D_t(u,v)
    \;\big|\; (u,v) \in \mathcal{M}_t
  \bigr\},
\end{equation}
where $\mathbf{K}$ is the camera intrinsic matrix and $\mathbf{T}^{\text{cam}}_t$ denotes the camera-to-base transformation obtained from the end-effector pose $\mathbf{T}^{\text{ee}}_t$ and a fixed hand-eye calibration.
Non-target masks are converted into coarse 3D boxes $\mathcal{B}_t = \{B_i \mid i \neq *\}$, which provide a compact occlusion model for viewpoint scoring.

\textbf{Target Completion under Registration Ambiguity.}
The partial cloud $\mathcal{P}^{\text{obs}}_t$ may not contain the task-relevant affordance, especially when the affordance is occluded by clutter or by the object itself. 
We therefore use SAM\,3D~\cite{sam3dteam2025sam3d3dfyimages} as a generative shape prior to imagine a complete target geometry from the masked RGB-D observation. The generated mesh is uniformly sampled to obtain a completed target cloud $\hat{\mathcal{P}}^{\text{comp}}_t$.
The imagined complete cloud is registered to the accumulated target observation using ICP with scale estimation~\cite{icp}. 
Under sparse or symmetric observations, multiple alignments can have similar registration fitness, leading to different possible affordance locations. 
Instead of selecting the best alignment, we retain the top-$K$ registration hypotheses. 
Each hypothesis $h_k$ defines a transformation $\mathbf{T}^{(k)}_{\text{reg}}$ and scale $\alpha^{(k)}$:
\begin{equation}
  \hat{\mathcal{P}}^{\text{comp},(k)}_t =
    \mathbf{T}^{(k)}_{\text{reg}}
    \bigl(\alpha^{(k)} \hat{\mathcal{P}}^{\text{comp}}_t\bigr),
    \quad k = 1, \ldots, K.
\end{equation}
Completion artifacts can also bias the alignment.
We address this by identifying ghost points using both 3D nearest-neighbor support and projected depth discrepancies to assess approximate conflicts with observed free space.
These points are penalized during coarse alignment and excluded before fine ICP, reducing their influence on the retained registrations. 
The aligned completions are thus used as spatial priors to be verified, rather than as exact reconstructions.

\textbf{Affordance Prediction.}
\label{sec:affordance}
We run GEAL~\cite{lu2024geal} on the completed cloud before registration, conditioned on $a^*$, to obtain a score $s_j\in[0,1]$ for each point $\mathbf{p}_j$.
To focus subsequent planning on the predicted functional region, we retain points above a task-specific threshold $\gamma$ and transform them under each registration:
\begin{equation}
\mathcal{A}^{(k)}_t
=
\left\{
\mathbf{p}^{(k)}_j
\;\middle|\;
\mathbf{p}_j\in\hat{\mathcal{P}}^{\mathrm{comp}}_t,\;
s_j\geq\gamma
\right\},
\label{eq:affordance_hypotheses}
\end{equation}
where
$\mathbf{p}^{(k)}_j
=\mathbf{T}^{(k)}_{\mathrm{reg}}(\alpha^{(k)}\mathbf{p}_j)$.
The scores remain unchanged across registrations, so the resulting sets describe alternative spatial placements of the same predicted affordance. 
These affordance-location hypotheses form the basis for the verification and disambiguation objectives below.

\subsection{Affordance-Targeted Next-Best-View Selection}
\label{sec:nbv}

Given the imagination-augmented scene representation, the view planner assesses whether the predicted affordance has been sufficiently verified and, if not, selects the next viewpoint to acquire further evidence.
Unlike reconstruction-driven NBV methods that maximize global object coverage, ATAP focuses on the thresholded affordance sets $\mathcal{A}^{(k)}_t$.
Its view selection balances two complementary objectives:
\emph{affordance verification}, which measures how much of the predicted affordance is supported by real depth observations, and
\emph{affordance-location disambiguation}, which aims to resolve ambiguity among the affordance locations proposed by competing registration hypotheses.

\textbf{Affordance-Location Hypotheses.}
The top-$K$ registrations from Sec.~\ref{sec:scene} induce $K$ possible spatial placements of the thresholded affordance set $\mathcal{A}^{(k)}_t$.
%
To account for their varying agreement with the accumulated target observations, we assign each hypothesis a registration reliability score:
\begin{equation}
\rho_k
=
-\bigl[(1-F_k)+\lambda_GG_k+\lambda_RR_k\bigr],
\label{eq:registration_reliability}
\end{equation}
where $F_k$ is the ICP correspondence inlier ratio, $R_k$ is the correspondence RMSE, and $G_k$ is the ghost-point ratio defined in Sec.~\ref{sec:scene}.
Thus, larger $\rho_k$ indicates better agreement.
We normalize these scores into hypothesis weights:
\begin{equation}
w_k
=
\frac{\exp(\rho_k/\sigma_\rho)}
{\sum_{j=1}^{K}\exp(\rho_j/\sigma_\rho)},
\label{eq:hypothesis_weights}
\end{equation}
where $\sigma_\rho=\max\{0.5(\max_j\rho_j-\min_j\rho_j),\,0.01\}$ is an adaptive temperature computed from their score range.
The affordance-location ambiguity is measured by
\begin{equation}
  H_t = -\sum_{k=1}^{K} w_k \log w_k .
\end{equation}
A high \(H_t\) indicates that several affordance placements remain plausible, whereas a low \(H_t\) indicates that one placement dominates.

\textbf{Affordance-Targeted Viewpoint Scoring.}
For each hypothesis $h_k$, we first partition the thresholded affordance set $\mathcal{A}^{(k)}_t$ into verified and unverified subsets using multi-view depth consistency. 
A point in $\mathcal{A}^{(k)}_t$ is verified if its valid image projection agrees with measured depth within $\epsilon_d$ in at least one acquired frame up to step $t$.
We denote these points by $\mathcal{S}^{(k)}_t$ and the
remaining points by $\mathcal{U}^{(k)}_t
=\mathcal{A}^{(k)}_t\setminus\mathcal{S}^{(k)}_t$.
Candidate camera positions are sampled on a spherical shell of radius $r=0.5$\,m centered at the target centroid, with the camera optical axis oriented toward the target center.
This target-centered sampling provides a common candidate set independent of any individual affordance-location hypothesis.
Workspace, inverse-kinematics, and collision checks are then applied to remove infeasible candidates, yielding the feasible set $\mathcal{V}_t$.
The candidates are subsequently ranked according to their expected affordance verification and disambiguation gains.
For each $v \in \mathcal{V}_t$, we estimate the fraction of the affordance set that is currently unverified and would become visible under each registration hypothesis:
\begin{equation}
  q^{(k)}_{\text{ver}}(v) =
    \frac{1}{|\mathcal{A}^{(k)}_t| + \epsilon}
    \sum_{p_j \in \mathcal{U}^{(k)}_t}
    \mathbf{1}\!\bigl[\text{visible}(p_j, v, \mathcal{B}_t)\bigr],
\end{equation}
where $\text{visible}(\cdot)$ checks the camera field of view and ray occlusion by the clutter boxes $\mathcal{B}_t$ and the target geometry, and $\epsilon$ is a small constant for numerical stability.
We compute the expected affordance verification gain by marginalizing over registration hypotheses:
\begin{equation}
  \bar{q}(v) = \sum_{k=1}^{K} w_k\, q^{(k)}_{\text{ver}}(v).
\end{equation}
This term rewards views exposing unverified affordance points, rather than merely increasing global object coverage.

Verification gain alone does not distinguish views that expose similar affordance fractions under competing hypotheses.
We therefore introduce a surrogate disambiguation gain based on the predicted affordance visibility under each hypothesis. 
Specifically, we use a binary observation model in which $q^{(k)}_{\mathrm{ver}}(v)$ serves as a surrogate likelihood of an affordance-consistent depth observation under $h_k$.
Conditioned on a positive or negative consistency outcome, the hypothesis weights are approximated as
\begin{equation}
      w^{+}_k(v) = \frac{w_k\, q^{(k)}_{\text{ver}}(v)}{\bar{q}(v)},
  \quad
  w^{-}_k(v) = \frac{w_k\,(1 - q^{(k)}_{\text{ver}}(v))}
                     {1 - \bar{q}(v)},
\end{equation}
where $\bar{q}(v)$ is clipped to $[\epsilon, 1{-}\epsilon]$ in the denominators to avoid degenerate updates. 
The expected posterior
entropy is
\begin{equation}
  \hat{H}_{t+1}(v) = \bar{q}(v)\,H(w^{+})
    + \bigl(1 - \bar{q}(v)\bigr)\,H(w^{-}),
\end{equation}
where $H(w) = -\sum_k w_k \log w_k$. The surrogate disambiguation gain is then
\begin{equation}
  \Delta H(v) = H_t - \hat{H}_{t+1}(v).
\end{equation}
A view yields a large $\Delta H(v)$ when different hypotheses predict different affordance visibility, making the view informative for resolving affordance-location ambiguity.

The final NBV score combines affordance verification, affordance-location disambiguation, and motion cost:
\begin{equation}
  v^* = \arg\max_{v \in \mathcal{V}_t}
  \Bigl[
    \underbrace{\Delta H(v)}_{\text{disambiguation}}
    + \underbrace{\bar{q}(v)}_{\text{verification}}
    - \underbrace{w_c\, c_{\text{move}}(v)}_{\text{motion cost}}
  \Bigr],
\end{equation}
where $c_{\mathrm{move}}(v)$ is a normalized weighted sum of Cartesian translational and angular displacements from the current camera pose $v_t$ to $v$.
When one hypothesis dominates, the potential entropy reduction is small, and view selection is driven primarily by affordance verification and motion cost.

\textbf{Termination and Closed-loop Update.}
Using the best-supported hypothesis $k^*$, we measure the
remaining need for verification by
\begin{equation}
r_t
=
\frac{|\mathcal{U}^{(k^*)}_t|}
{|\mathcal{A}^{(k^*)}_t|+\epsilon}.
\label{eq:unverified_ratio}
\end{equation}
Coverage alone may leave competing locations unresolved,
while low entropy does not ensure that the functional region
has been observed.
We therefore terminate the NBV loop when
\begin{equation}
\bigl(r_t<\tau_r\;\land\;H_t<\tau_H\bigr)
\quad\lor\quad t\geq T,
\label{eq:joint_termination}
\end{equation}
where $t$ counts viewpoint adjustments after the initial
observation and $T$ is the maximum budget.

After executing \(v^*\), the robot acquires a new RGB-D observation and updates the accumulated target cloud and the occluder boxes $\mathcal{B}_t$. 
The perception pipeline is then re-executed to update hypothesis weights and the revised verified/unverified affordance partition. 
This observe--imagine--verify loop repeats until the termination criterion is met, after which the system proceeds to TOG execution.


\subsection{Task-oriented Grasp Execution}
\label{sec:grasp}

After the NBV loop terminates, we execute the final grasp using the verified affordance under the current best-supported hypothesis $k^*$.
Let $\mathcal{A}^{\text{ver}}_t = \mathcal{S}^{(k^*)}_t$ denote the affordance points that have been corroborated by real observations. 
We run AnyGrasp~\cite{anygrasp} on the accumulated target point cloud to generate 6-DoF grasp candidates. 
Each grasp candidate $g$ is scored by combining grasp quality and affordance alignment:
\begin{equation}
  Q(g) =
    \lambda_1 s_{\text{AG}}(g)
    + \lambda_2 \phi_{\text{cen}}(g)
    + \lambda_3 \phi_{\text{near}}(g),
\end{equation}
where $s_{\text{AG}}(g) \in [0,1]$ is the grasp quality returned by AnyGrasp \cite{anygrasp}.
The two affordance-alignment terms are defined as
\[
     \phi_{\text{cen}}(g) =
    \exp\!\biggl(-\frac{d^2_{\text{cen}}(g)}{2\sigma_c^2}\biggr),
  \quad
  \phi_{\text{near}}(g) =
    \exp\!\biggl(-\frac{d^2_{\text{near}}(g)}{2\sigma_n^2}\biggr), 
\]
where $d_{\text{cen}}(g)$ is the distance from the grasp center to the centroid of $\mathcal{A}^{\text{ver}}_t$, and $d_{\text{near}}(g)$ is the distance to the nearest verified affordance point. 
These terms bias grasp selection toward the verified task-relevant region, while $s_{\text{AG}}$ preserves geometric grasp stability. 
Among the candidates that pass inverse kinematics and collision checking, the grasp with the highest $Q(g)$ is executed.

\section{Experiments}
We evaluate ATAP in simulation and on a real robot to examine
whether affordance-targeted active perception improves both
functional grasp success and viewpoint efficiency under occlusion.
We further study how performance changes with initial affordance occlusion, whether multi-hypothesis planning benefits ambiguous scenes, and whether these advantages transfer to real-world clutter.

\subsection{Experiment Setup}

\begin{figure}[t]
  \centering
  \includegraphics[width=0.325\linewidth]{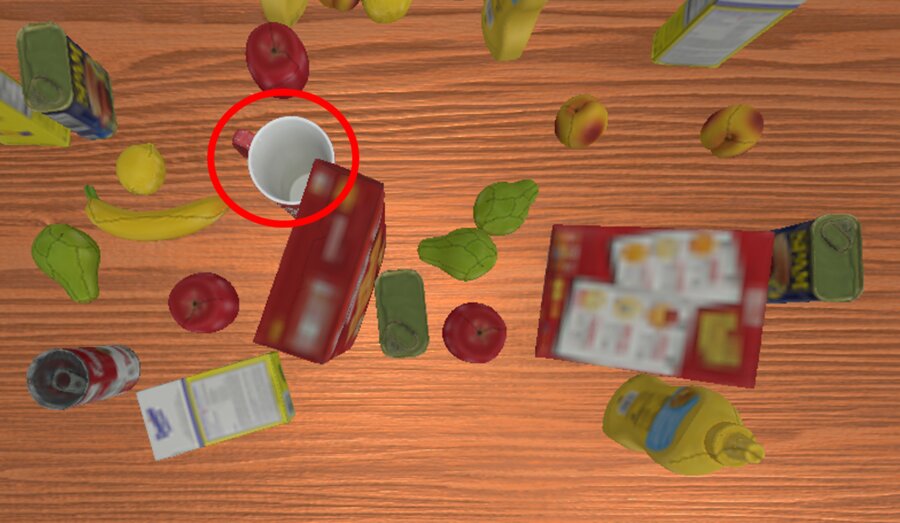}
  \hfill
  \includegraphics[width=0.325\linewidth]{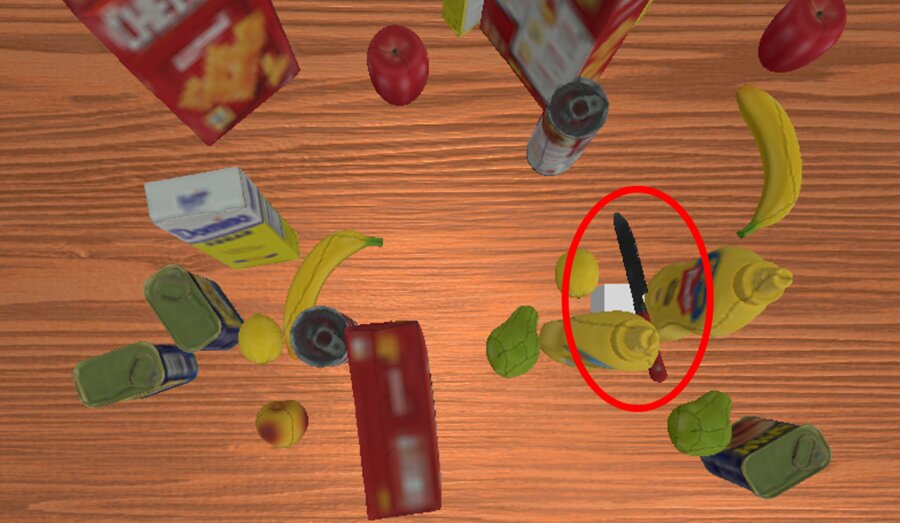}
  \hfill
  \includegraphics[width=0.325\linewidth]{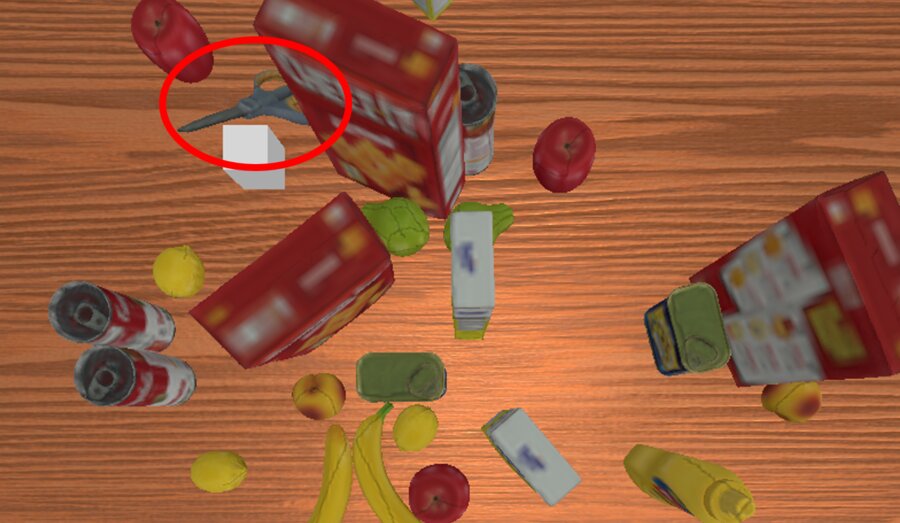}
  \makebox[0.325\linewidth]{\small (a) 
  }%
  \hfill
  \makebox[0.325\linewidth]{\small (b) 
  }%
  \hfill
  \makebox[0.325\linewidth]{\small (c) 
  }%
    \caption{Simulation benchmark scenes on ManiSkill3.
    (a) Mug-handle grasping for pouring.
    (b) Knife-handle grasping for cutting.
    (c) Scissors finger-loop grasping for cutting.
    All three tasks use a top-down camera as the initial viewpoint.
    Red circles mark the target objects, whose affordance regions are occluded by surrounding clutter or the object’s own pose.}
  \label{fig:sim_scenes}
\end{figure}

\textbf{Simulation Environment.}
All simulation experiments are conducted on the ManiSkill3 platform~\cite{ManiSkill3} using a 7-DoF Franka Emika Panda manipulator with a parallel-jaw gripper and a wrist-mounted RGB-D camera ($1080{\times}720$, $90^\circ$ FoV). 
The robot operates on a $0.8{\times}1.0$\,m tabletop workspace populated with randomly placed household objects. 
We evaluate all methods with three tasks that span different affordance geometries:
(1) pouring with a mug (handle grasp),
(2) cutting with a knife (handle grasp), and
(3) cutting with scissors (finger-loop grasp), as illustrated in Fig.~\ref{fig:sim_scenes}.
Following~\cite{ostg}, thin objects such as knives and scissors are elevated 5\,cm above the table surface to ensure graspability with a parallel-jaw gripper. 
For each task, we sample 100 random scenes with varying object poses and occlusion configurations.

\begin{table*}[t]
\centering
\caption{
Simulation results across three task-oriented grasping tasks on ManiSkill3.
GS and FGS denote Grasp Success Rate and Functional Grasp Success Rate (\%), respectively.
Steps denotes the average number of viewpoint adjustments after the initial observation.
Best FGS are shown in \textbf{bold}.
}
\label{tab:sim_results}
\setlength{\tabcolsep}{5.5pt}
\begin{tabular}{l | c c c | c c c | c c c | c c c}
\toprule
& \multicolumn{3}{c|}{Mug (Handle)}
& \multicolumn{3}{c|}{Knife (Handle)}
& \multicolumn{3}{c|}{Scissors (Loop)}
& \multicolumn{3}{c}{Average} \\
\cmidrule(lr){2-4}
\cmidrule(lr){5-7}
\cmidrule(lr){8-10}
\cmidrule(lr){11-13}
Method
& GS$\uparrow$ & FGS$\uparrow$ & Steps$\downarrow$
& GS$\uparrow$ & FGS$\uparrow$ & Steps$\downarrow$
& GS$\uparrow$ & FGS$\uparrow$ & Steps$\downarrow$
& GS$\uparrow$ & FGS$\uparrow$ & Steps$\downarrow$ \\
\midrule

\multicolumn{13}{l}{\textit{Part A: Comparison with Baselines}} \\
\midrule

AffordGrasp
& 28 & 27 & 0
& 36 & 21 & 0
& 29 & 19 & 0
& 31.0 & 22.3 & 0 \\

Recon.\ NBV
& 53 & 53 & 4.31
& 65 & 59 & 4.10
& 55 & 50 & 3.19
& 57.7 & 54.0 & 3.87 \\

Semantic NBV
& 34 & 33 & 0.49
& 73 & 48 & 0.37
& 45 & 39 & 0.32
& 50.7 & 40.0 & 0.39 \\

\midrule
\multicolumn{13}{l}{\textit{Part B: Ablation Studies}} \\
\midrule

w/o Active Perception
& 45 & 42 & 0
& 67 & 56 & 0
& 34 & 27 & 0
& 48.7 & 41.7 & 0 \\

w/o Multi-hypothesis
& 66 & 63 & 2.94
& 60 & 56 & 0.85
& 60 & 60 & 2.04
& 62.0 & 59.7 & 1.94 \\

\midrule

\textbf{ATAP (Ours)}
& 80 & \textbf{68} & 2.43
& 69 & \textbf{65} & 0.63
& 69 & \textbf{66} & 1.90
& 72.7 & \textbf{66.3} & 1.65 \\

\bottomrule
\end{tabular}
\end{table*}

\textbf{Implementation Details.}
At $t{=}0$, we use GPT-4o \cite{openai2024gpt4ocard} once to parse the language instruction to determine the target object $O^*$ and affordance type $a^*$. 
Grounding DINO~\cite{GroundingDino} and SAM2~\cite{ravi2024sam2} provide instance-level target segmentation.
SAM\,3D~\cite{sam3dteam2025sam3d3dfyimages} performs shape completion from the RGB-D observation and target mask, and GEAL~\cite{lu2024geal} predicts per-point affordance scores conditioned on $a^*$.
ATAP retains $K{=}3$ registration hypotheses. The registration reliability score uses a ghost penalty weight of $0.3$ and an ICP residual weight of $0.5\,\mathrm{m}^{-1}$. The viewpoint objective uses $w_c{=}0.3$, with $|\mathcal{V}|{=}128$ candidate viewpoints and a maximum viewpoint budget of $T{=}5$. The depth-consistency tolerance is set to $\epsilon_d{=}0.005$\,m. The verification and ambiguity thresholds are set to $\tau_r{=}\text{0.2}$ and 
$\tau_H{=}\text{0.1}$, respectively.
Grasp candidates are generated by AnyGrasp~\cite{anygrasp}
and ranked using $\lambda_1{=}0.6$, $\lambda_2{=}0.3$, and
$\lambda_3{=}0.1$, with distance scales
$\sigma_c{=}0.05$\,m and $\sigma_n{=}0.01$\,m.
We use affordance thresholds $\gamma=0.4$, $0.2$, and $0.05$ for mug, knife, and scissors, respectively.
All parameters are fixed before evaluation and remain unchanged across scenes of the same task.

\textbf{Baselines.}
We compare ATAP with fixed-viewpoint and active perception baselines.
\textbf{AffordGrasp}~\cite{affordgrasp} performs open-vocabulary task-oriented grasping directly from the same initial observation used by ATAP, without subsequent viewpoint adjustment.
\textbf{\textit{Reconstruction-based NBV (Recon. NBV)}} adopts the volumetric information gain of Breyer et al.~\cite{breyer2022} to reconstruct the target object, after which the same affordance prediction and grasp generation pipeline as ATAP is applied.
\textbf{\textit{Semantic-guided NBV}} adopts the occlusion-semantic viewpoint optimization of APeG~\cite{apeg} and terminates when AffordGrasp~\cite{affordgrasp} successfully grounds the affordance region for grasp execution.
All active-perception methods use the same maximum viewpoint
budget of $T=5$.


\textbf{Metrics.}
We evaluate all methods with three metrics.
\textit{Grasp Success Rate} (\textbf{GS}) measures whether the target object is lifted and held stably for at least 3\,s.
\textit{Functional Grasp Success Rate} (\textbf{FGS}) additionally requires the grasp contact points to lie within 1.5 cm of the ground-truth affordance region.
It is our primary metric for TOG evaluation.
\textit{Average NBV Steps} (\textbf{Steps}) counts the mean number of viewpoint adjustments across all trials.

\subsection{Simulation Results}

\label{sec:sim_res}

Table~\ref{tab:sim_results} summarizes the results across three TOG tasks. 
Part~A compares ATAP with baselines, while Part~B evaluates ablated variants of the proposed pipeline.

\textbf{Comparison with fixed-viewpoint baselines.}
AffordGrasp yields substantially lower FGS than ATAP across all tasks, highlighting the limitation of single-view TOG when the task-relevant region is occluded. By actively acquiring additional observations to verify the predicted affordance, ATAP provides more reliable evidence for TOG.


\textbf{Comparison with active perception baselines.}
Recon.~NBV improves grasping under occlusion by actively recovering target geometry, but its reconstruction-driven objective requires more viewpoint adjustments. In contrast, ATAP achieves higher FGS with 57\% fewer viewpoint adjustments, showing that directing perception toward the task-relevant affordance uses the view budget more effectively than reconstructing the entire target. 
Semantic NBV requires fewer viewpoint adjustments but yields much lower FGS, indicating that improving target-level semantic visibility does not necessarily expose the functional region required for TOG.

\textbf{Ablation studies.}
Removing active perception reduces FGS, showing that imagined affordance geometry alone cannot replace verification from real observations. Nevertheless, this variant still outperforms the fixed-view AffordGrasp baseline, which suggests that shape imagination can partially compensate for missing visual evidence by hypothesizing plausible hidden affordance geometry. However, these hypotheses may be inaccurate under severe occlusion, motivating active verification before grasp execution. 
Restricting ATAP to a single registration hypothesis reduces FGS while increasing viewpoint adjustments, showing the benefit of explicitly maintaining alternative affordance locations under ambiguous observations.


\begin{figure}[t]
\centering
\includegraphics[width=0.8\linewidth]{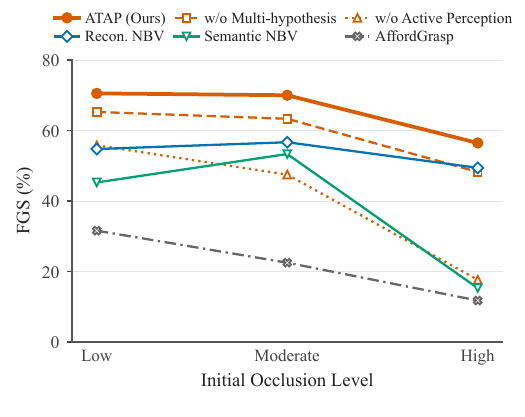}
\caption{
\textbf{Robustness to initial occlusion.}
Functional grasp success rate (FGS) across low, moderate, and high initial occlusion levels.
}
\label{fig:occlusion}
\end{figure}

\textbf{Robustness to initial affordance occlusion.}
Following TARGO's same-view visibility comparison~\cite{targo}, we quantify initial affordance occlusion offline using ground-truth affordance geometry.
With the camera and target poses fixed, let $V_{\mathrm{full}}$ and $V_{\mathrm{ref}}$ denote the visible affordance surface in the complete scene and an affordance-only reference, respectively. We define $o_{\mathrm{aff}}=1-V_{\mathrm{full}}/V_{\mathrm{ref}}$, where a larger value indicates stronger occlusion, and divide the 300 scenes into low ($0$--$30\%$), moderate ($30$--$60\%$), and high ($60$--$100\%$) occlusion levels.
As shown in Fig.~\ref{fig:occlusion}, ATAP achieves the highest FGS across all three levels and remains comparatively stable as initial affordance visibility decreases.
In contrast, the fixed-view and semantic-guided baselines degrade sharply under severe occlusion, while Recon.~NBV is less sensitive to the initial view but remains below ATAP.
These results show that affordance-targeted verification is particularly beneficial when the task-relevant region is heavily occluded in the initial view.

\begin{table}[t]
\centering
\caption{
Multi-hypothesis ablation on 85 scenes with high
affordance occlusion.
}
\label{tab:hard_ambiguity}
\begin{tabular}{lcc}
\toprule
Method & FGS $\uparrow$ & Steps $\downarrow$ \\
\midrule
w/o Multi-hypothesis
& 48.24\% & 3.14 \\
ATAP (Ours)
& \textbf{56.47\%} & \textbf{2.51} \\
\bottomrule
\end{tabular}
\end{table}


\begin{table}[t]
\centering
\caption{Real-world evaluation results over 20 scenes per task.}
\label{tab:real}
\begin{tabular}{l | c c c | c c c}
\toprule
& \multicolumn{3}{c|}{Mug (Handle)}
& \multicolumn{3}{c}{Scissors (Loop)} \\
Method & GS$\uparrow$ & FGS$\uparrow$ & Steps$\downarrow$
       & GS$\uparrow$ & FGS$\uparrow$ & Steps$\downarrow$ \\
\midrule
AffordGrasp$^\dagger$ & 15/20 & 4/20 & --
                       & 12/20 & 3/20 & -- \\
\textbf{ATAP (Ours)}  & 15/20 & \textbf{13/20} & 1.25
                       & 13/20 & \textbf{12/20} & 1.3 \\
\bottomrule
\multicolumn{7}{l}{\small $\dagger$: Fixed side-view observation.} \\
\end{tabular}
\end{table}

\begin{figure}[t]
\centering
\includegraphics[width=0.65\linewidth]{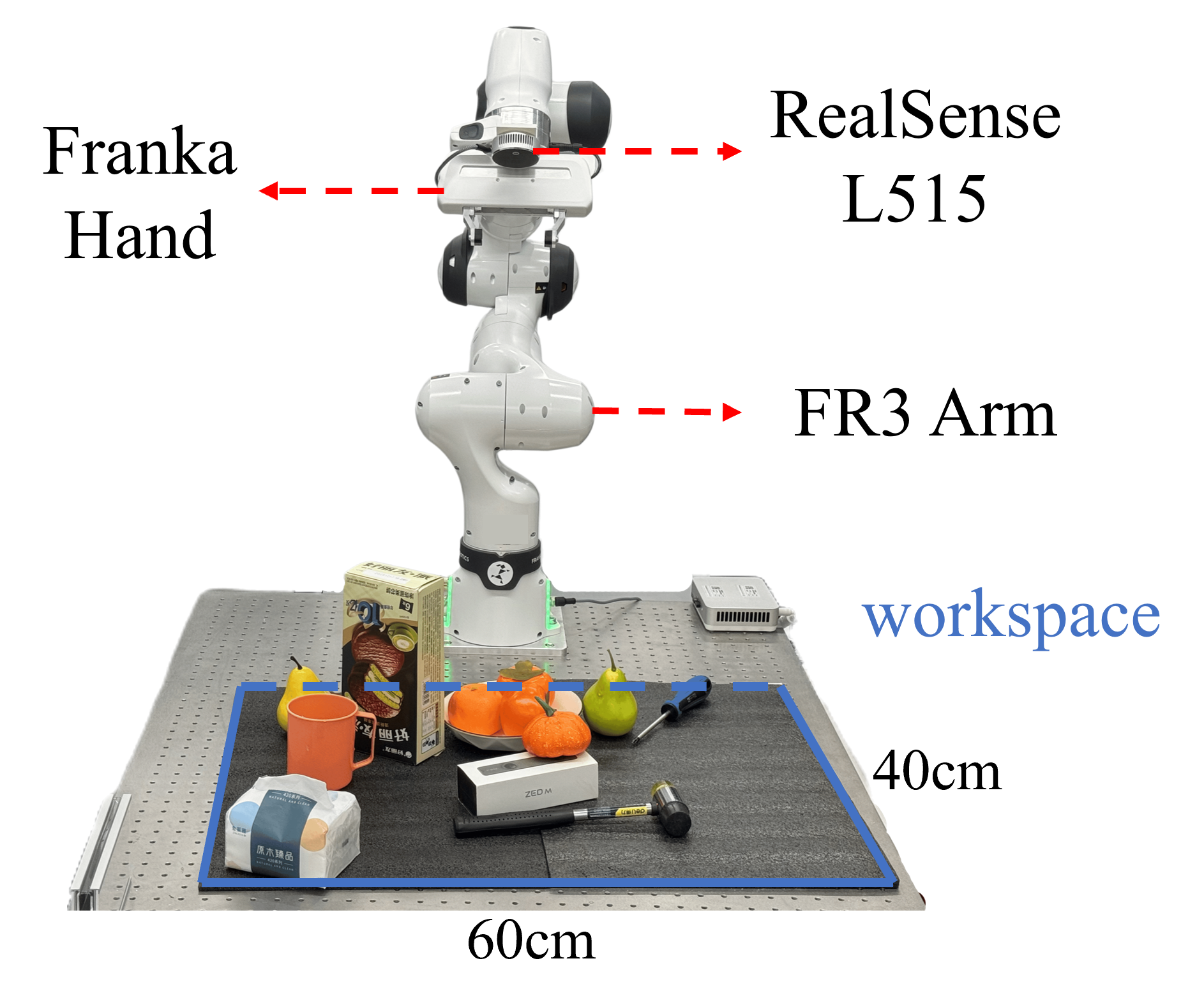}%
\caption{Real-robot setup with an FR3 arm, Franka Hand gripper, and wrist-mounted RealSense L515 RGB-D camera.}
\label{fig:real}
\end{figure}

\begin{figure}[t]
    \centering
    \begin{subfigure}[b]{0.445\columnwidth}
        \centering
        \includegraphics[width=\linewidth]{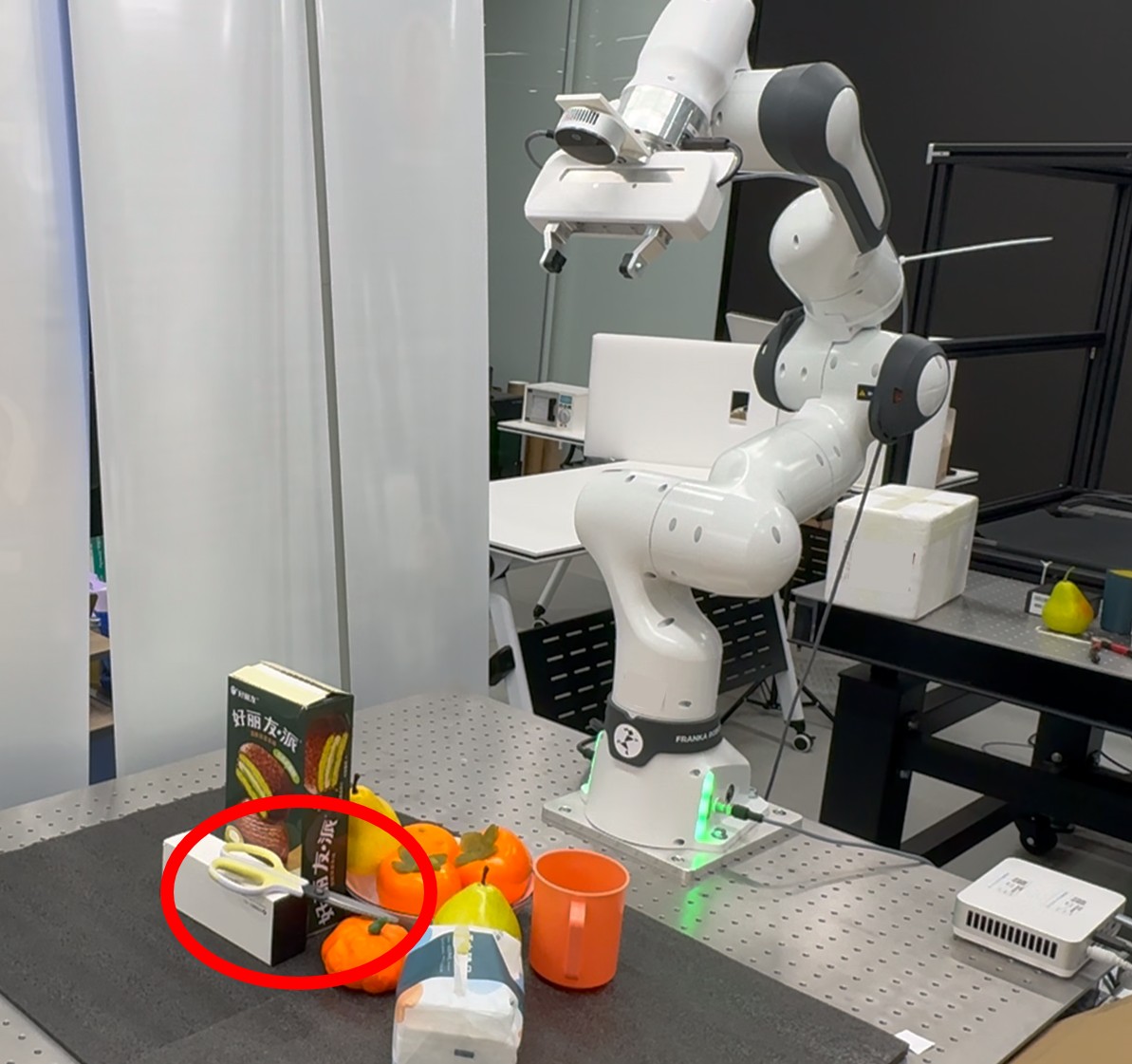}
        \caption{Initial View}
        \label{fig:real_init_3rd}
    \end{subfigure}
    \hfill
    \begin{subfigure}[b]{0.45\columnwidth}
        \centering
        \includegraphics[width=\linewidth]{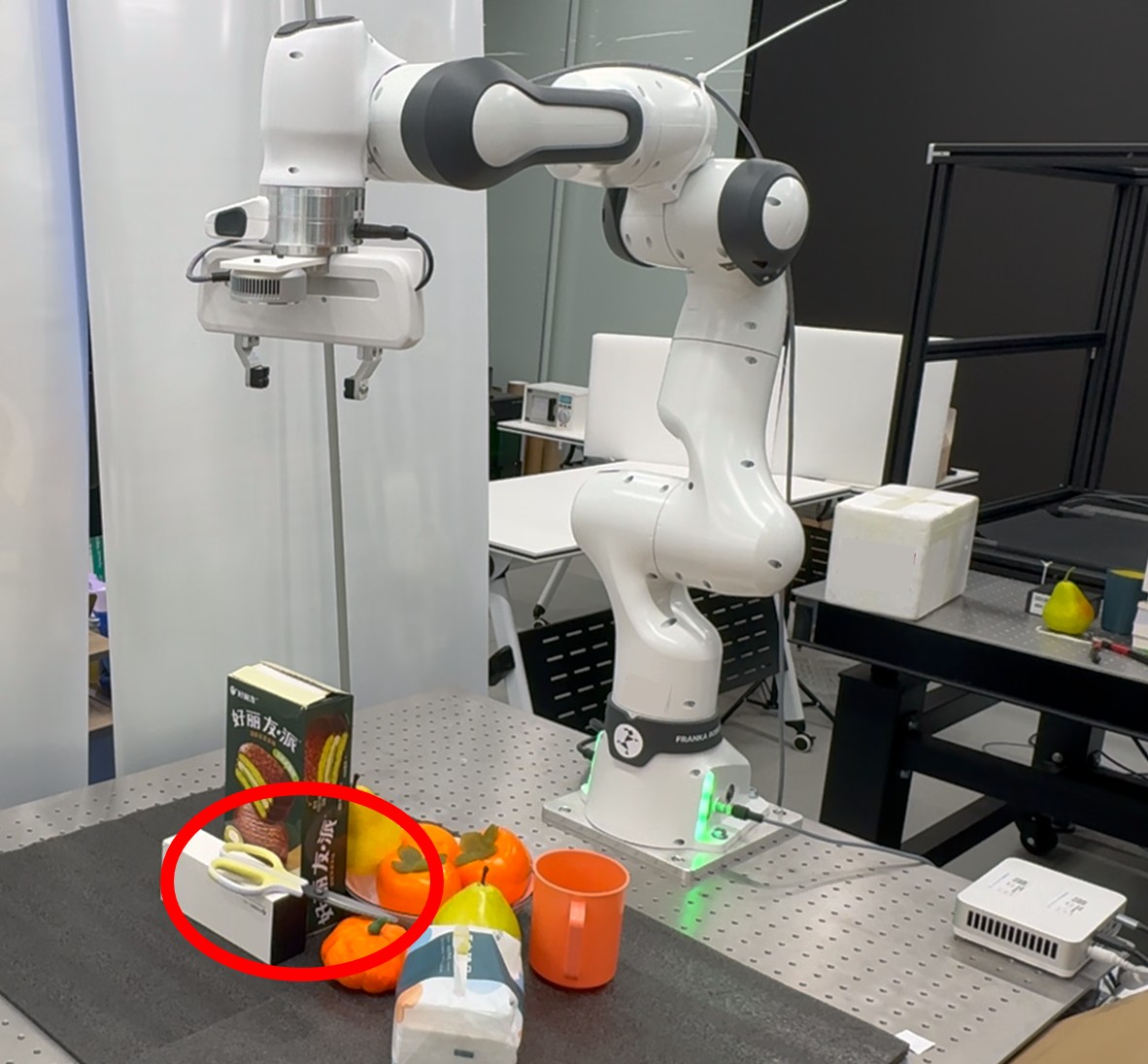}
        \caption{Active View}
        \label{fig:real_nbv_3rd}
    \end{subfigure}
    
    \vspace{2pt}
    
    \begin{subfigure}[b]{0.45\columnwidth}
        \centering
        \includegraphics[width=\linewidth]{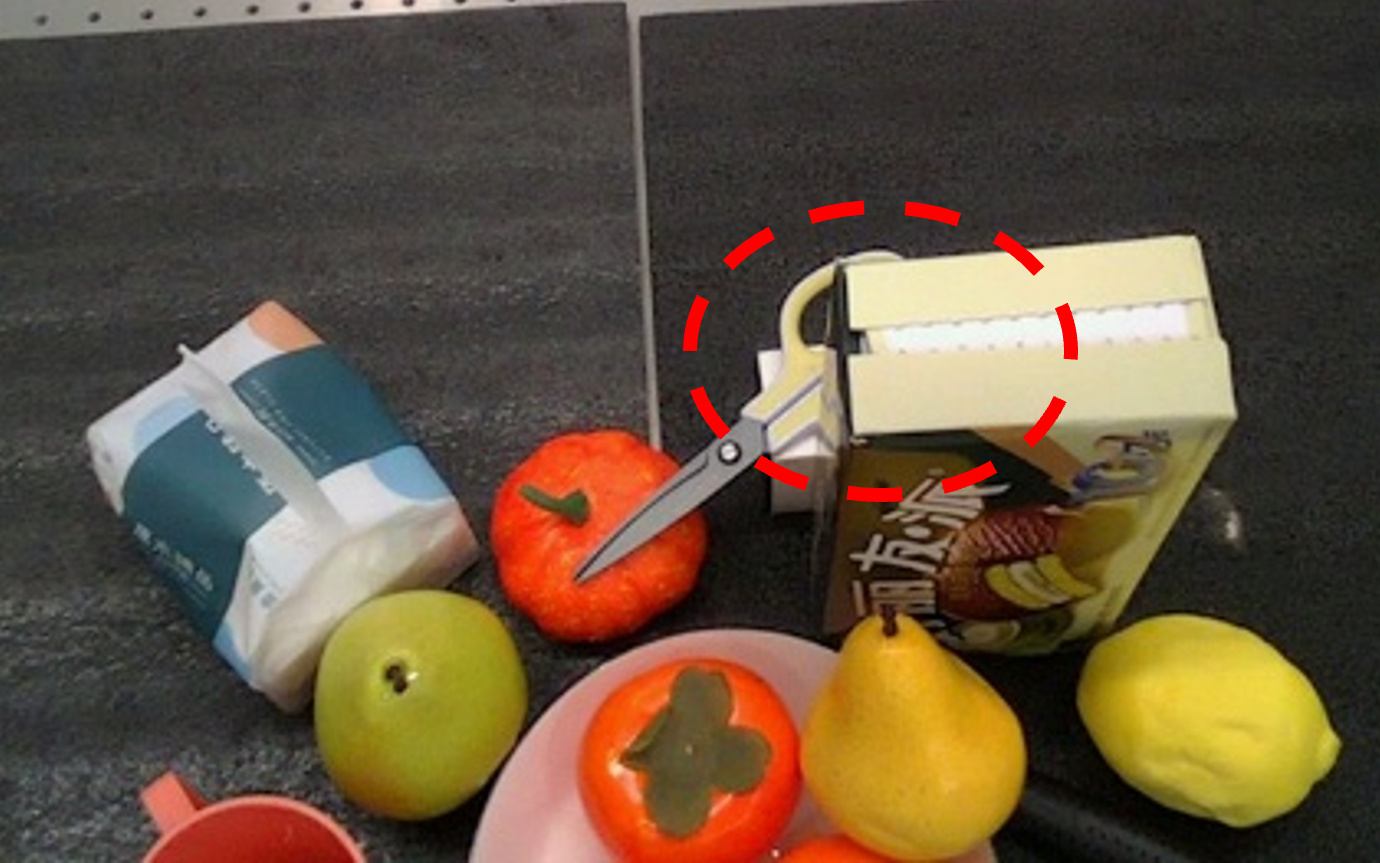}
        \caption{Initial Observation}
        \label{fig:real_init_wrist}
    \end{subfigure}
    \hfill
    \begin{subfigure}[b]{0.45\columnwidth}
        \centering
        \includegraphics[width=\linewidth]{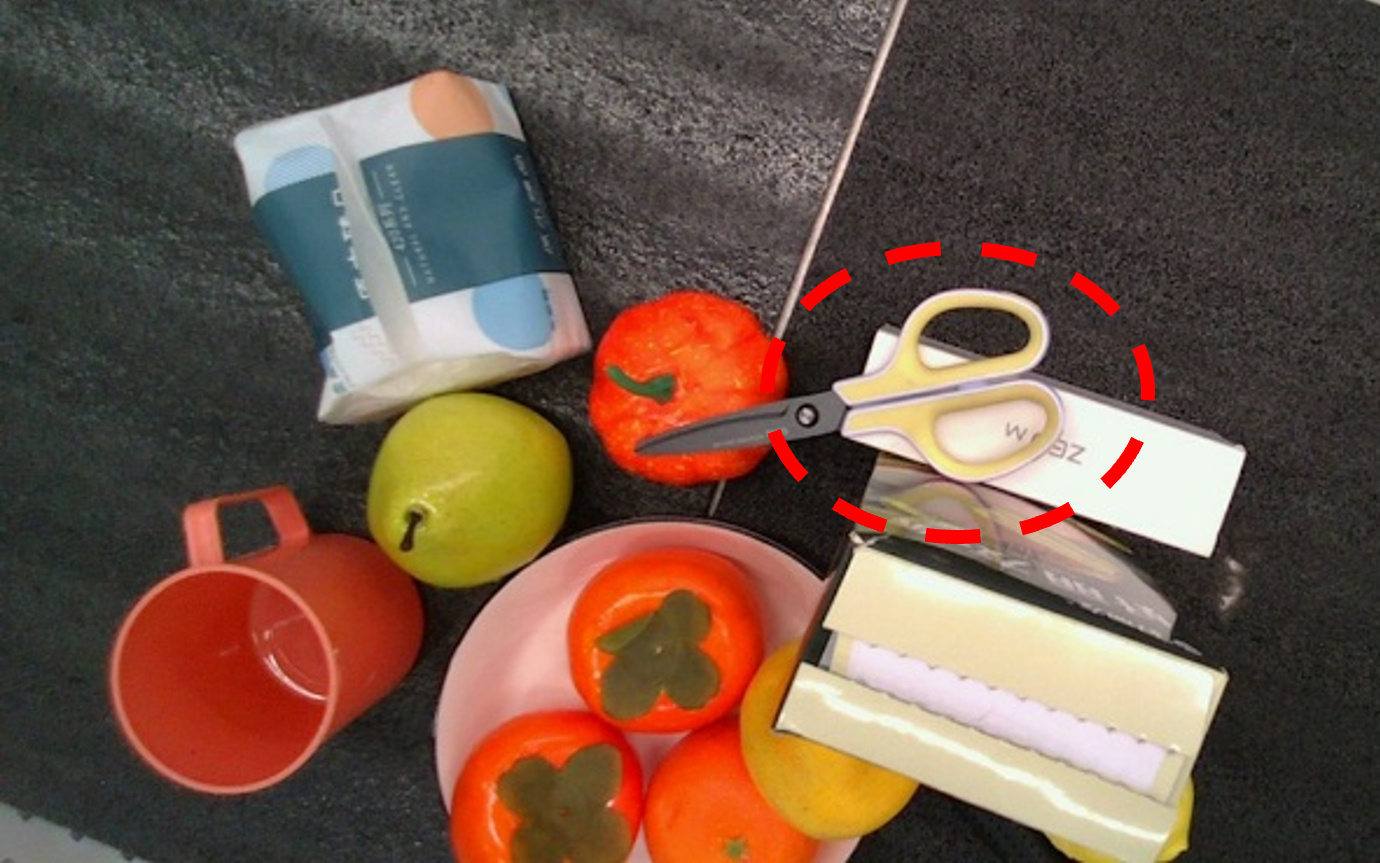}
        \caption{Active Observation}
        \label{fig:real_nbv_wrist}
    \end{subfigure}
    
    \caption{Real-robot scissors grasping trial. Top row: third-person view of the robot at the initial pose (a) and after active viewpoint adjustment (b). Bottom row: corresponding wrist-camera observations. The red dashed circle marks the target affordance region (finger loop). The finger loop is occluded from the initial view. ATAP selects an active viewpoint that exposes the affordance region, enabling a successful TOG.}
    \label{fig:real_robot_qualitative}
\end{figure}
\textbf{Multi-hypothesis ablation under high affordance occlusion.}
Severe affordance occlusion can leave the target registration ambiguous,
motivating the retention of alternative affordance-location hypotheses.
We evaluated this design on 85 scenes with initial
$o_{\mathrm{aff}} \geq 60\%$.
Specifically, we compared ATAP with \emph{w/o Multi-hypothesis},
which selects viewpoints using only the best-supported registration
at each iteration.
ATAP achieved higher FGS with fewer viewpoint adjustments
(Table~\ref{tab:hard_ambiguity}), which suggests that explicitly favoring views that distinguish competing affordance locations improves both grasp reliability and view efficiency under severe occlusion.

\subsection{Real-World Validation}

We deploy ATAP on a 7-DoF Franka Research 3 (FR3) arm equipped with the Franka Hand parallel-jaw gripper and a wrist-mounted Intel RealSense L515 RGB-D camera in the eye-in-hand configuration (Fig.~\ref{fig:real}). 
We compare ATAP against AffordGrasp~\cite{affordgrasp} from a fixed side-view observation on two tasks (mug handle and scissors finger loop), and each task is evaluated over 20 scenes with different clutter arrangements.
Both methods start from the same fixed side-view observation.

As shown in Table~\ref{tab:real}, ATAP achieves substantially higher FGS than AffordGrasp with comparable GS, indicating that fixed-view observations often reach the target but miss the task-relevant affordance. 
By actively verifying the predicted affordance with few viewpoint adjustments, ATAP exposes the functional region for stable task-oriented grasp execution. 
Fig.~\ref{fig:real_robot_qualitative} shows a representative scissors trial, where the initially occluded finger loop is exposed after active viewpoint adjustment, enabling a successful grasp on the verified affordance.

\begin{figure}[t]
\centering
\includegraphics[width=1.0\linewidth]{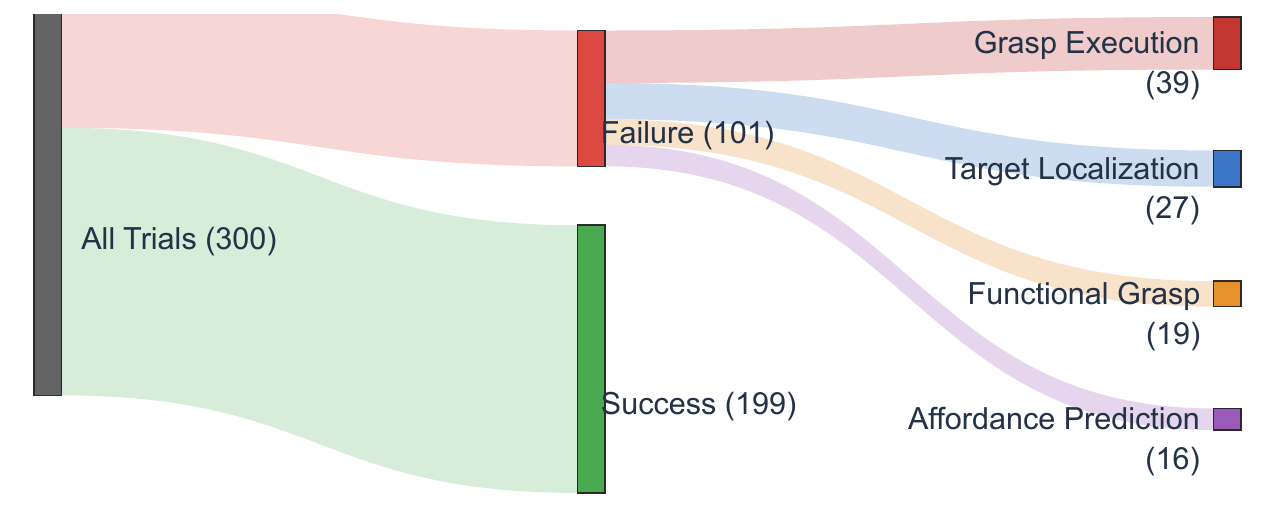}
\caption{
Failure flow analysis of ATAP over 300 simulation scenes.}
\label{fig:failure}
\end{figure}

\subsection{Failure Case Analysis}
To diagnose the remaining limitations of ATAP, we assign each
unsuccessful simulation scene to its primary failure stage, as
summarized in Fig.~\ref{fig:failure}.
\emph{Target localization} failures mainly occur under severe
occlusion, revealing the dependence of the current framework on reliable initial grounding and segmentation.
\emph{Affordance prediction} failures typically originate from
inaccurate shape completion, where erroneous imagined geometry
leads to incorrect affordance hypotheses and can subsequently
misguide view selection.
\emph{Functional grasp selection} fails when the affordance is
correctly localized but no stable grasp is available on the
functional region, exposing a mismatch between task relevance
and local graspability.
Finally, \emph{grasp execution} is the largest failure category, where otherwise valid affordance-aligned grasps become unstable during physical execution.
Overall, these failures suggest that further gains require tighter coupling between active perception, affordance reasoning, and graspability-aware execution.

\section{CONCLUSIONS}

We presented ATAP, a closed-loop active perception framework for task-oriented grasping under affordance occlusion.
ATAP follows a hypothesize-then-verify strategy that uses generative shape completion to hypothesize hidden affordance regions and actively verifies them through targeted depth observations. 
By combining affordance verification with affordance-location disambiguation, ATAP directs viewpoint planning toward task-relevant regions instead of exhaustive object reconstruction. 
Simulation and real-world experiments show that ATAP improves functional grasp success while requiring fewer viewpoint adjustments than reconstruction-based active perception.

\addtolength{\textheight}{-2cm}   

\bibliographystyle{IEEEtran}
\bibliography{IEEEabrv,root}

\begin{thebibliography}{10}
\providecommand{\url}[1]{#1}
\csname url@samestyle\endcsname
\providecommand{\newblock}{\relax}
\providecommand{\bibinfo}[2]{#2}
\providecommand{\BIBentrySTDinterwordspacing}{\spaceskip=0pt\relax}
\providecommand{\BIBentryALTinterwordstretchfactor}{4}
\providecommand{\BIBentryALTinterwordspacing}{\spaceskip=\fontdimen2\font plus
\BIBentryALTinterwordstretchfactor\fontdimen3\font minus \fontdimen4\font\relax}
\providecommand{\BIBforeignlanguage}[2]{{%
\expandafter\ifx\csname l@#1\endcsname\relax
\typeout{** WARNING: IEEEtran.bst: No hyphenation pattern has been}%
\typeout{** loaded for the language `#1'. Using the pattern for}%
\typeout{** the default language instead.}%
\else
\language=\csname l@#1\endcsname
\fi
#2}}
\providecommand{\BIBdecl}{\relax}
\BIBdecl

\bibitem{catgrasp}
B.~Wen \emph{et~al.}, ``Catgrasp: Learning category-level task-relevant grasping in clutter from simulation,'' in \emph{2022 International Conference on Robotics and Automation (ICRA)}, 2022, pp. 6401--6408.

\bibitem{tog}
A.~Murali \emph{et~al.}, ``Same object, different grasps: Data and semantic knowledge for task-oriented grasping,'' in \emph{Conference on robot learning}.\hskip 1em plus 0.5em minus 0.4em\relax PMLR, 2021, pp. 1540--1557.

\bibitem{foundationgrasp}
C.~Tang \emph{et~al.}, ``Foundationgrasp: Generalizable task-oriented grasping with foundation models,'' \emph{IEEE Transactions on Automation Science and Engineering}, vol.~22, pp. 12\,418--12\,435, 2025.

\bibitem{graspnet-1b}
H.-S. Fang \emph{et~al.}, ``Graspnet-1billion: A large-scale benchmark for general object grasping,'' in \emph{Proceedings of the IEEE/CVF Conference on Computer Vision and Pattern Recognition (CVPR)}, June 2020.

\bibitem{anygrasp}
H.-S. Fang \emph{et~al.}, ``Anygrasp: Robust and efficient grasp perception in spatial and temporal domains,'' \emph{IEEE Transactions on Robotics}, vol.~39, no.~5, pp. 3929--3945, 2023.

\bibitem{graspgpt}
C.~Tang \emph{et~al.}, ``Graspgpt: Leveraging semantic knowledge from a large language model for task-oriented grasping,'' \emph{IEEE Robotics and Automation Letters}, vol.~8, no.~11, pp. 7551--7558, 2023.

\bibitem{rtagrasp}
W.~Dong \emph{et~al.}, ``Rtagrasp: Learning task-oriented grasping from human videos via retrieval, transfer, and alignment,'' in \emph{2025 IEEE International Conference on Robotics and Automation (ICRA)}, 2025, pp. 1--7.

\bibitem{affordgrasp}
Y.~Tang \emph{et~al.}, ``Affordgrasp: In-context affordance reasoning for open-vocabulary task-oriented grasping in clutter,'' in \emph{2025 IEEE/RSJ International Conference on Intelligent Robots and Systems (IROS)}, 2025, pp. 9433--9439.

\bibitem{ovalgrasp}
\BIBentryALTinterwordspacing
E.~Tong \emph{et~al.}, ``Oval-grasp: Open-vocabulary affordance localization for task oriented grasping,'' 2025. [Online]. Available: \url{https://arxiv.org/abs/2511.20841}
\BIBentrySTDinterwordspacing

\bibitem{thinkgrasp}
Y.~Qian \emph{et~al.}, ``Thinkgrasp: A vision-language system for strategic part grasping in clutter,'' in \emph{8th Annual Conference on Robot Learning}, 2024.

\bibitem{breyer2022}
M.~Breyer \emph{et~al.}, ``Closed-loop next-best-view planning for target-driven grasping,'' in \emph{2022 IEEE/RSJ International Conference on Intelligent Robots and Systems (IROS)}, 2022, pp. 1411--1416.

\bibitem{ace-nbv}
X.~Zhang \emph{et~al.}, ``Affordance-driven next-best-view planning for robotic grasping,'' in \emph{Proceedings of The 7th Conference on Robot Learning}, ser. Proceedings of Machine Learning Research, J.~Tan \emph{et~al.}, Eds., vol. 229.\hskip 1em plus 0.5em minus 0.4em\relax PMLR, 06--09 Nov 2023, pp. 2849--2862.

\bibitem{apeg}
Y.~Dai \emph{et~al.}, ``Active-perceptive language-oriented grasp policy for heavily cluttered scenes,'' \emph{IEEE Robotics and Automation Letters}, vol.~10, no.~11, pp. 11\,094--11\,101, 2025.

\bibitem{viso-grasp}
Y.~Shi \emph{et~al.}, ``Viso-grasp: Vision-language informed spatial object-centric 6-dof active view planning and grasping in clutter and invisibility,'' in \emph{2025 IEEE/RSJ International Conference on Intelligent Robots and Systems (IROS)}, 2025, pp. 14\,931--14\,938.

\bibitem{contactgrasp-net}
M.~Sundermeyer \emph{et~al.}, ``Contact-graspnet: Efficient 6-dof grasp generation in cluttered scenes,'' in \emph{2021 IEEE International Conference on Robotics and Automation (ICRA)}, 2021, pp. 13\,438--13\,444.

\bibitem{Gater}
M.~Sun and Y.~Gao, ``Gater: Learning grasp-action-target embeddings and relations for task-specific grasping,'' \emph{IEEE Robotics and Automation Letters}, vol.~7, no.~1, pp. 618--625, 2022.

\bibitem{partAffordance}
Y.~Song \emph{et~al.}, ``Learning 6-dof fine-grained grasp detection based on part affordance grounding,'' \emph{IEEE Transactions on Automation Science and Engineering}, vol.~22, pp. 15\,200--15\,214, 2025.

\bibitem{shapegrasp}
S.~Li \emph{et~al.}, ``Shapegrasp: Zero-shot task-oriented grasping with large language models through geometric decomposition,'' in \emph{2024 IEEE/RSJ International Conference on Intelligent Robots and Systems (IROS)}, 2024, pp. 10\,527--10\,534.

\bibitem{ostg}
A.-L. Wang \emph{et~al.}, ``Task-oriented 6-dof grasp pose detection in clutters,'' in \emph{2025 IEEE International Conference on Robotics and Automation (ICRA)}, 2025, pp. 5692--5698.

\bibitem{tosc}
W.~Wu \emph{et~al.}, ``{TOSC}: Task-oriented shape completion for open-world dexterous grasp generation from partial point clouds,'' in \emph{Proceedings of the AAAI Conference on Artificial Intelligence}, vol.~40, no.~13, 2026, pp. 10\,781--10\,789.

\bibitem{active-ngf}
H.~Ma \emph{et~al.}, ``Active perception for grasp detection via neural graspness field,'' in \emph{Advances in Neural Information Processing Systems}, A.~Globerson \emph{et~al.}, Eds., vol.~37.\hskip 1em plus 0.5em minus 0.4em\relax Curran Associates, Inc., 2024, pp. 38\,122--38\,141.

\bibitem{ap-vlm}
\BIBentryALTinterwordspacing
V.~Sripada \emph{et~al.}, ``Scene exploration by vision-language models,'' 2025. [Online]. Available: \url{https://arxiv.org/abs/2409.17641}
\BIBentrySTDinterwordspacing

\bibitem{openai2024gpt4ocard}
\BIBentryALTinterwordspacing
OpenAI \emph{et~al.}, ``Gpt-4o system card,'' 2024. [Online]. Available: \url{https://arxiv.org/abs/2410.21276}
\BIBentrySTDinterwordspacing

\bibitem{GroundingDino}
S.~Liu \emph{et~al.}, ``Grounding dino: Marrying dino with grounded pre-training for open-set object detection,'' in \emph{Computer Vision -- ECCV 2024}, A.~Leonardis \emph{et~al.}, Eds.\hskip 1em plus 0.5em minus 0.4em\relax Cham: Springer Nature Switzerland, 2025, pp. 38--55.

\bibitem{ravi2024sam2}
N.~Ravi \emph{et~al.}, ``Sam 2: Segment anything in images and videos,'' in \emph{International Conference on Learning Representations}, Y.~Yue \emph{et~al.}, Eds., vol. 2025, 2025, pp. 28\,085--28\,128.

\bibitem{sam3dteam2025sam3d3dfyimages}
X.~Chen \emph{et~al.}, ``Sam 3d: 3dfy anything in images,'' in \emph{Proceedings of the IEEE/CVF Conference on Computer Vision and Pattern Recognition (CVPR)}, June 2026, pp. 7220--7232.

\bibitem{icp}
\BIBentryALTinterwordspacing
P.~J. Besl and N.~D. McKay, ``A method for registration of 3-d shapes,'' \emph{IEEE Trans. Pattern Anal. Mach. Intell.}, vol.~14, no.~2, p. 239–256, Feb. 1992. [Online]. Available: \url{https://doi.org/10.1109/34.121791}
\BIBentrySTDinterwordspacing

\bibitem{lu2024geal}
D.~Lu \emph{et~al.}, ``Geal: Generalizable 3d affordance learning with cross-modal consistency,'' in \emph{Proceedings of the IEEE/CVF Conference on Computer Vision and Pattern Recognition (CVPR)}, June 2025, pp. 1680--1690.

\bibitem{ManiSkill3}
S.~Tao \emph{et~al.}, ``Maniskill3: Gpu parallelized robotics simulation and rendering for generalizable embodied ai,'' \emph{Robotics: Science and Systems}, 2025.

\bibitem{targo}
Y.~Xia \emph{et~al.}, ``Targo: Benchmarking target-driven object grasping under occlusions,'' \emph{arXiv preprint arXiv:2407.06168}, 2024.

\end{thebibliography}

\end{document}